\documentclass[letterpaper, 10 pt, conference]{ieeeconf}  

\IEEEoverridecommandlockouts                              

\usepackage{graphics} 
\usepackage{epsfig} 
\usepackage{mathptmx} 
\usepackage{times} 
\usepackage{amsmath} 
\usepackage{amssymb}  
\usepackage{graphicx}
\usepackage{booktabs}
\usepackage{comment}
\usepackage{float} 
\usepackage[table,xcdraw]{xcolor}
\usepackage{subcaption}
\usepackage[colorlinks, linkcolor=blue]{hyper ref}

\usepackage[symbol]{footmisc}

\newcommand{\ie}{{\em i.e.}}
\newcommand{\eg}{{\em e.g.}}

\newcommand{\etc}{{\em etc.}}

\usepackage[capitalize]{cleveref}
\DeclareMathOperator*{\argmin}{\arg\!\min}

\title{\LARGE \bf
Scaling Vision-based End-to-End Autonomous Driving with Multi-View Attention Learning
}

\author{Yi Xiao$^{1}$, Felipe Codevilla$^{2}$, Diego Porres$^{1}$ and Antonio M. L\'opez$^{1}$
\thanks{$^{1}$Yi Xiao, Diego Porres and Antonio M. L\'opez are with Department of Computer Science, Computer Vision Center (CVC), Universitat Autònoma de Barcelona (UAB), Spain. {\tt\small \{yxiao, dporres, antonio\}@cvc.uab.cat}}%
\thanks{$^{2}$Felipe Codevilla is with Montreal Institute for Learning Algorithms (MILA), Montreal, Canada. {\tt\small felipe.alcm@gmail.com}}%
}

\begin{document}

\maketitle
\thispagestyle{empty}
\pagestyle{empty}

\begin{abstract}
On end-to-end driving, human driving demonstrations are used to train perception-based driving models by imitation learning. This process is supervised on vehicle signals (\eg, steering angle, acceleration) but does not require extra costly supervision (human labeling of sensor data).
As a representative of such vision-based end-to-end driving models, CILRS is commonly used as a baseline to compare with new driving models.
So far, some latest models achieve better performance than CILRS by using expensive sensor suites and/or by using large amounts of human-labeled data for training. Given the difference in performance, one may think that it is not worth pursuing vision-based pure end-to-end driving. 
However, we argue that this approach still has great value and potential considering cost and maintenance. In this paper, we present CIL++, which improves on CILRS by both processing higher-resolution images using a human-inspired HFOV as an inductive bias and incorporating a proper attention mechanism. CIL++ achieves competitive performance compared to models which are more costly to develop.
We propose to replace CILRS with CIL++ as a strong vision-based pure end-to-end driving baseline supervised by only vehicle signals and trained by conditional imitation learning.
\end{abstract}

\section{INTRODUCTION}
End-to-end autonomous driving (EtE-AD) refers to deep models trained to process sensor data for performing maneuvers that imitate human (expert) driving \cite{Tampuu:2022}. Broadly, we can classify these models according to their training supervision requirements. Some models only require vehicle signals ({\eg}, steering angle, acceleration) as supervision. Eventually, they can be directly trained from millions of human driving experiences. Other models also require costly human-based sensor data labeling as supervision ({\eg}, pixel/voxel-wise semantic labels or 2D/3D object bounding boxes). In fact, these models follow a kind of hybrid approach leveraging from pure EtE-AD and traditional AD pipelines \cite{Grigorescu:2020, Yurtsever:2020}.  

Despite the appealing idea of developing vision-based pure EtE-AD models, their progress has mostly stalled, giving space to hybrid models supervised by a significantly large amount of labeled sensor data \cite{Muller:2018,Liang:2018,Chen:2019,Zhang:2021,Chitta:2021,Prakash:2021,Chen:2022,Shao:2022,Hu:2022,Wu:2022}, or by privileged information from the driving environment as required by reinforcement learning \cite{Zhang:2021,Toromanoff:2020}. This situation may be due to the apparent lack of scalability of pure EtE-AD models raised in a few works \cite{De:2019,Spencer:2021}; including \cite{Codevilla:2019}, where the model known as CILRS is proposed. CILRS was developed in suboptimal conditions: limited driving episodes based on a single and rather deterministic expert driver, very low-resolution images depicting a relatively narrow horizontal field of view (on-board images roughly display a single lane), and without applying any attention mechanism. Overall, this gives rise to poor performance in newer benchmarks, eventually misleading non-expert readers and new practitioners in the field regarding the potential of vision-based pure EtE-AD. 

Since having strong baselines is important to avoid illusory gains when developing new models, we present\footnote[1]{CIL++ model and code are publicly released on \href{https://github.com/yixiao1/CILv2_multiview}{GitHub}
} \emph{CIL++}, a strong vision-based pure EtE-AD model trained by conditional imitation learning, {\ie}, CILRS. We improve on CILRS key limitations, rising the performance of CIL++ to be on par with top-performing hybrid methods. First, we drastically increase the image view by using a horizontal field of view (HFOV) similar to human drivers, which runs on $180^\circ-220^\circ$. Further, we propose  a visual transformer-based architecture \cite{Vaswani:2017} which acts as a mid-level attention mechanism for these views, allowing CIL++ to associate feature map patches (tokens) across different views. These changes allow CIL++ to perform at the expert level on the CARLA \emph{NoCrash} metrics. Moreover, CIL++ is the first vision-based pure EtE-AD model capable of obtaining competitive results on complex CARLA towns.

\begin{figure*}[!t]
  \centering
  \includegraphics[width=1.0\linewidth]{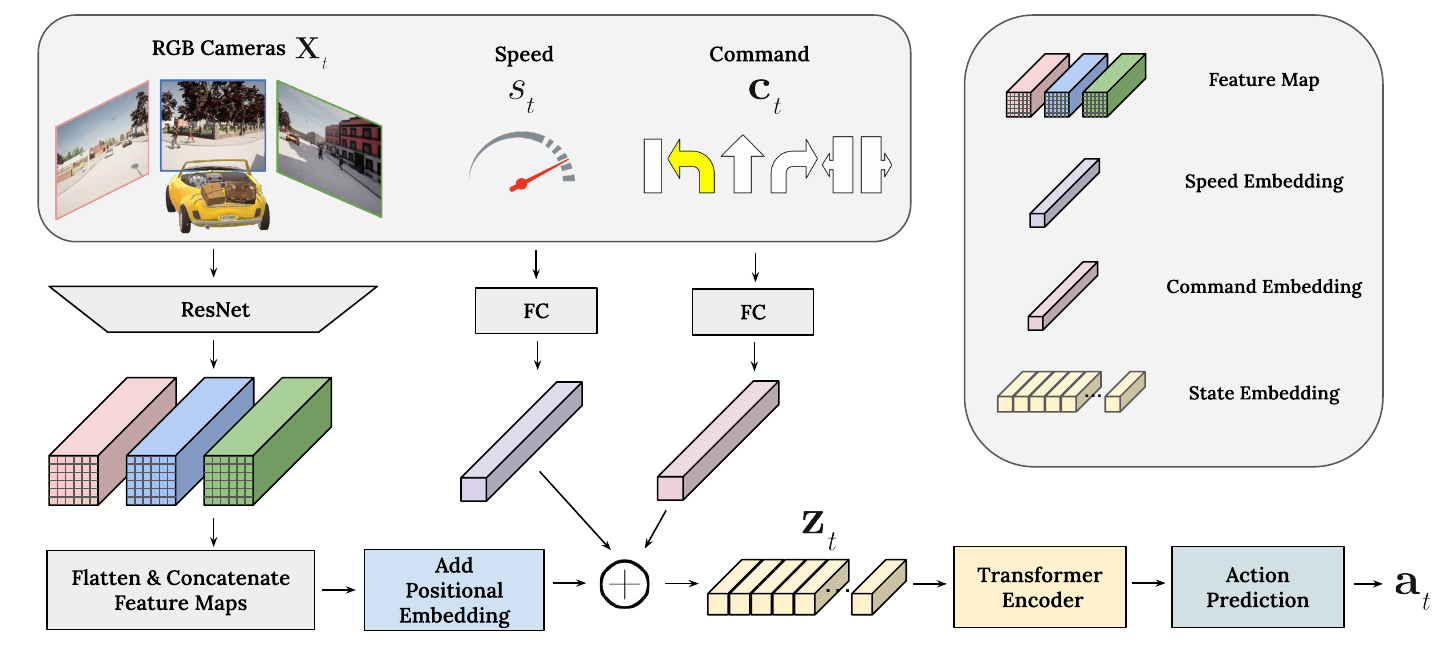}
  \caption{CIL++'s architecture: our model is mainly comprised of three parts: state embedding, transformer encoder, and action prediction. Specifically, the input observation consists of multi-view RGB images $\mathbf{X}_{t}$, ego-vehicle's speed $s_t$, and high-level navigation command $\mathbf{c}_{t}$. The output is action $\mathbf{a}_{t}$, which is directly applied to maneuver the ego-vehicle. Since $\mathbf{a}_{t}$ consists of the ego-vehicle's steering angle and acceleration, CIL++ is a vision-based pure EtE-AD model. Note that these action components are automatically read and associated with the captured images during data collection.}
  \label{fig:architecture}
\end{figure*}

\section{Related Work}
Learning a driving policy from experts, instead of handcrafting it, is a really appealing idea. Accordingly, in EtE-AD, a deep model is trained by imitation learning, which has become an active research topic \cite{Tampuu:2022}. Pioneering works following this approach are \cite{Pomerleau:1989, LeCun:2005, Bojarski:2016, Bojarski:2017}. However, it was the public release of the CARLA simulator \cite{Dosovitskiy:2017}, together with a vision-based pure EtE-AD model trained and tested on CARLA \cite{Codevilla:2018}, that attracted great attention to this paradigm  \cite{Muller:2018,Bansal:2019,Bewley:2019,Codevilla:2019,Xiao:2020,Zhang:2021,Chitta:2021,Prakash:2021,Chen:2022,Shao:2022,Hu:2022,Wu:2022}.

Basically, we can find two ways of approaching EtE-AD according to the output of the underlying deep model. On the one hand, different proposals output waypoints on bird-eye-view (BeV) coordinates \cite{Muller:2018,Bansal:2019,Chen:2019,Chitta:2021,Prakash:2021,Chen:2022,Shao:2022}. These are then used by a controller for providing the proper steering and acceleration in driving. Note that generating BeVs involves 3D knowledge of the scene. On the other hand, other proposals directly output the ego-vehicle signals (steering, acceleration) \cite{Codevilla:2018,Pan:2018,Liang:2018,Bewley:2019,Codevilla:2019,Xiao:2020,Zhang:2021,Hu:2022}, which act on the vehicle either directly or after some signal-stabilizer filtering ({\eg}, using a PID). Both approaches were combined, {\eg}, in \cite{Wu:2022}, where a branch of the EtE-AD model is used to predict waypoints and another to predict ego-vehicle signals, being the model output a combination of both, weighted according to the perceived road curvature.

Another important difference among EtE-AD models is the type of required supervision for their training. For instance, some models require semantic segmentation labels at training time \cite{Chen:2019,Chitta:2021,Hu:2022}, sometimes together with object bounding boxes (BBs) \cite{Prakash:2021}, or BBs and HD maps \cite{Bansal:2019}. Other models are only supervised by ego-vehicle available signals \cite{Pomerleau:1989,LeCun:2005,Bojarski:2016,Bojarski:2017, Codevilla:2019}, thus no human-labeled sensor data is required. EtE-AD models can also fuse multi-modal sensor data such as RGB image and depth from either LiDAR \cite{Chen:2022,Prakash:2021,Shao:2022} or monocular depth estimation \cite{Xiao:2020}. 


In recent literature, vision-based pure EtE-AD models such as CILRS \cite{Codevilla:2018} are clearly outperformed by those using additional supervision (pixel-wise semantic labels, object BBs, {\etc}). For instance, the currently top-performing hybrid model, termed as MILE \cite{Hu:2022}, relies on a training procedure requiring $\approx3$M images labeled for semantic segmentation. Another top-performing hybrid model, NEAT \cite{Chitta:2021}, requires $\approx130$K of those. In addition, Roach \cite{Zhang:2021} leverages the teacher/student mechanism. First, a \emph{teacher} model (Roach RL) is trained using environment-privileged information and reinforcement learning. Then, it supervises the training of a \emph{student} model (Roach IL) by applying imitation learning. In this paper, we refer to the student model as RIM.

CILRS was developed under completely different conditions than we have today: only using single-lane towns in CARLA simulator (Town01 and Town02), using very low-resolution images depicting a narrow horizontal field of view (roughly, one-lane-width views), without including any attention mechanism, and relying on the default CARLA's expert driver, which follows handcrafted rules. Overall, this drives it to perform poorly on newer benchmarks.

Our model, CIL++, is a direct successor of CILRS, aiming at bringing back the competitiveness of vision-based pure EtE-AD. For training and testing CIL++, we use the multi-town setting available from CARLA 0.9.13. Moreover, as in recent works \cite{Hu:2022}, to collect training data (images and ego-vehicle signals), we use an expert \cite{Zhang:2021} with better driving performance than the CARLA's default one. Note that, in simulation, this expert plays the role of a human driver in the real world, who would drive to collect on-board data. In addition, we use a wider horizontal field of view (HFOV=$180^\circ$), in the range of human drivers. This kind of inductive bias is crucial to avoid injecting undesired causal confusion when training the model. For instance, while collecting driving episodes for training, the ego-vehicle may be stopped at a red traffic light because the expert driver has access to the privileged information of the environment. However, this red light may not even be captured by the on-board camera due to a narrow HFOV. This was observed in CILRS which was using HFOV=$100^\circ$. In fact, using wide HFOVs has become a common practice to develop EtE-AD models ({\eg}, \cite{Zhang:2021,Chitta:2021,Prakash:2021,Shao:2022,Hu:2022}). As we would do in the real world to minimize image distortion, we use three forward-facing cameras with HFOV=$60^\circ$ each. Finally, in order to jointly consider the image content from the three cameras (views), we propose to use Transformer \cite{Vaswani:2017} which acts as a mid-level attention mechanism for these views. All these improvements over CILRS make CIL++ competitive.

Like CILRS, we use the current ego-vehicle speed and high-level navigation commands as input signals to the model. However, unlike CILRS, we also consider left/right lane changes as possible command values at testing time. Note that some supervised methods such as MILE use as input the road shape pattern to be expected according to the current position of the ego-vehicle, instead of processing a high-level navigation command. Given a curve, this pattern can be more or less curved according to the corresponding lane curvature. In CILRS and CIL++, high-level commands ({\eg}, \emph{continue} in this lane) are equivalent to those from a navigation system for global planning. In fact, such navigation commands together with the ego-vehicle speed are the only information that CIL++ uses beyond the multi-view images. This is in contrast with other models such as NEAT \cite{Chitta:2021}, which use explicit traffic light detection at testing time.

As we will see, even CIL++ is not using supervision at all, it outperforms RIM and is quite on par with MILE. 

\section{METHOD}

\subsection{Problem Setup}
CIL++ is trained by imitation learning, which we formalize as follows. Expert demonstrators (drivers) produce an action $\mathbf{a}_{i}$ (ego-vehicle maneuver) when encountering an observation $\mathcal{O}_i$, (sensor data, signals) given the expert policy $\pi^{\star}(\mathcal{O}_{i})$ (driving skills, attitude, {\etc}). The basic idea behind imitation learning is to train an agent (here CIL++) that mimics an expert by using these observations.

Prior to training an agent, we need to collect a dataset comprised of observation/action pairs $\mathcal{D} = \left\{(\mathbf{o}_{i}, \mathbf{a}_{i})\right\}_{i=1}^N$ generated by the expert. This dataset is in turn used to train a policy $\pi_{\theta}(\mathbf{o}_{i})$  which approximates the expert policy. The general imitation learning objective is then

\begin{equation} \label{eq:target}
  \argmin_{\theta} \mathbb{E}_{(\mathbf{o}_{i}, \mathbf{a}_{i})\sim \mathcal{D}} \left[ \mathcal{L}(\pi_{\theta}(\mathbf{o}_{i}), \mathbf{a}_{i}) \right] \enspace .
\end{equation}

During testing time, we assume that only the trained policy  $\pi_{\theta}(\mathbf{o}_{i})$ will be used and no expert will be available. 

\subsection{Architecture}

Figure \ref{fig:architecture} overviews CIL++'s architecture. Our model is mainly comprised of three parts: state embedding, transformer encoder, and action prediction module.

\subsubsection{State Embedding}
At time $t$, the current state $\mathbf{X}_t$ consists of a set of images from the left, central, and right cameras $\mathbf{X}_{t}=\{\mathbf{x}_{l,t}, \mathbf{x}_{c,t}, \mathbf{x}_{r,t} \}$, the ego vehicle's forward speed $s_t\in\mathbb{R}$, and a high-level navigation command $\mathbf{c}_t\in\mathbb{R}^{k}$ which is encoded as a one-hot vector. 

As discussed in \cite{Alexey:2021}, the lack of inductive biases makes transformer models require more data to achieve good performance. In order to possess the inherent properties of CNNs (\ie, exploiting locality and translation equivariance), as well as leveraging the attention mechanism of transformers, we propose to adapt the hybrid model suggested in \cite{Alexey:2021} to our case. At time $t$, each image $\mathbf{x}_{v,t}\in \mathbb{R}^{W\times H\times 3}$ from the multi-view camera setting is processed by a share-weight ResNet34 \cite{He:2016}, pre-trained on ImageNet \cite{Deng:2009}. Then, for each view $v$, we take the resulting feature map $\mathbf{f}_{v,t}\in\mathbb{R}^{w\times h\times c}$ from the last convolutional layer of ResNet34, where $w\times h$ is the spatial size and $c$ indicates the feature dimension. Each feature map is then flattened along the spatial dimensions, resulting in $P\times c$ tokens, where $P=w*h$ is the number of spatial features per image. Since we will set our cameras to a resolution of $W\times H=300\times300$ pixels, for each one we obtain  $P=100$ patches with $c=512$ from the ResNet34 backbone. 

Since we use $|\mathbf{X}_{t}|=3$ views (left, central, and right cameras), we take the flattened patches for each view, and tokenize them as the whole sequence with length $S=|\mathbf{X}_t|*w*h$ for further feeding  into the transformer model. To provide the positional information for each token, we apply the standard learnable 1D positional embedding $\mathbf{p}\in \mathbb{R}^{S\times c}$ as done in \cite{Alexey:2021}, which is added directly to the token. 

The forward speed $s_t$ and command $\mathbf{c}_t$ are linearly projected to $\mathbb{R}^{c}$ using a fully connected layer. The resulting state embedding $\mathbf{z}_t$ is obtained by the addition of these input embeddings. 
Formally, we define the state encoder of the current state $\mathcal{X}_t=(\mathbf{X}_t, s_t, \mathbf{c}_t)$, parameterized by $\theta$, as
\begin{equation}\label{eq:encoder}
    e_{\theta}: \mathbb{R}^{|\mathbf{X}_{t}|\times W\times H\times3}\times\mathbb{R}\times \mathbb{R}^k \rightarrow \mathbb{R}^{S \times c} \enspace .
\end{equation}

In CIL \cite{Codevilla:2018} and CILRS \cite{Codevilla:2019}, $\mathbf{c}_{t}$ is treated as a switcher (condition) to trigger different MLP branches for predicting $\mathbf{a}_{t}$. Here we treat $\mathbf{c}_{t}$ as an input signal to be later processed by a transformer. This is because we have not observed obvious differences between these two approaches while treating $\mathbf{c}_{t}$ as input signal simplifies the training.  

\subsubsection{Attention Learning}
To naturally associate intra-view and inter-view information, we adopt the attention mechanism of transformers \cite{Vaswani:2017}. We expect it to be effective to learn the mutual relevance between distant image patches (tokens), helping our model to associate feature map patches across views ({\ie}, coming from visual information to the left, center, and right of the ego-vehicle). 

Over the embedded space $\mathbf{Z}_{t}$, we learn a scenario embedding using a transformer encoder that consists of $L$ multi-head attention layers. Each one includes Multi-headed Self-Attention (MHSA) \cite{Vaswani:2017}, layer normalization (LN) \cite{Ba:2016}, and feed-forward MLP blocks.  
The final output is a linear projection of the concatenated output of each attention head, which is then fed into the action prediction module. We use $L=4$ layers, with $4$ heads each. The hidden dimension $D$ of the transformer layer is set to be equal to the ResNet output dimension, {\ie}, $D=512$.

\subsubsection{Action Prediction}
The output of the transformer encoder with a size of $S\times c$ is average-pooled and fed into an MLP. The MLP consists of three fully connected layers (FC) with ReLu non-linearity applied between each FC. The final output action $\mathbf{a}_t\in\mathbb{R}^2$ comprises of the steering angle and acceleration (brake/throttle), {\ie}, $\mathbf{a}_{t} = (a_{\text{s},t}, a_{\text{acc}, t})$. 

\subsection{Loss Function}
At time $t$, given a predicted action $\mathbf{a}_{t}$ and a ground truth action $\hat{\mathbf{a}}_{t}$, we define the training loss as:
\begin{equation}\label{eq:loss}
 \mathcal{L}(\mathbf{a}_t, \hat{\mathbf{a}}_t) = \lambda_{\text{acc}}\lVert a_{\text{acc},t}-\hat{a}_{\text{acc,t}}\rVert_{1} + \lambda_{\text{s}} \lVert a_{\text{s},t}-\hat{a}_{\text{s},t} \rVert_{1} \enspace ,
\end{equation}
where $\lVert\cdot\rVert_{1}$ is the $L_1$ distance, $\lambda_{acc}$ and $\lambda_s$ indicate the weights given to the acceleration and steering angle loss, respectively. In our case, we consider steering angle and acceleration to be both in the range of $[-1, 1]$. Negative values of the acceleration correspond to braking, while positive ones to throttle. The weights are set to $\lambda_{acc} = \lambda_{s} = 0.5$. 
 
In CILRS \cite{Codevilla:2019}, speed prediction regularization is applied in the training loss to avoid the inertia problem caused by the overwhelming probability of ego staying static in the training data. We do not observe this problem in our case, thus the speed prediction branch is not applied in our setting. Our result suggests that a simple $L_1$ loss is able to provide compelling performance, even in a new town.

\begin{figure}[t]
  \centering
  \includegraphics[width=1.0\linewidth]{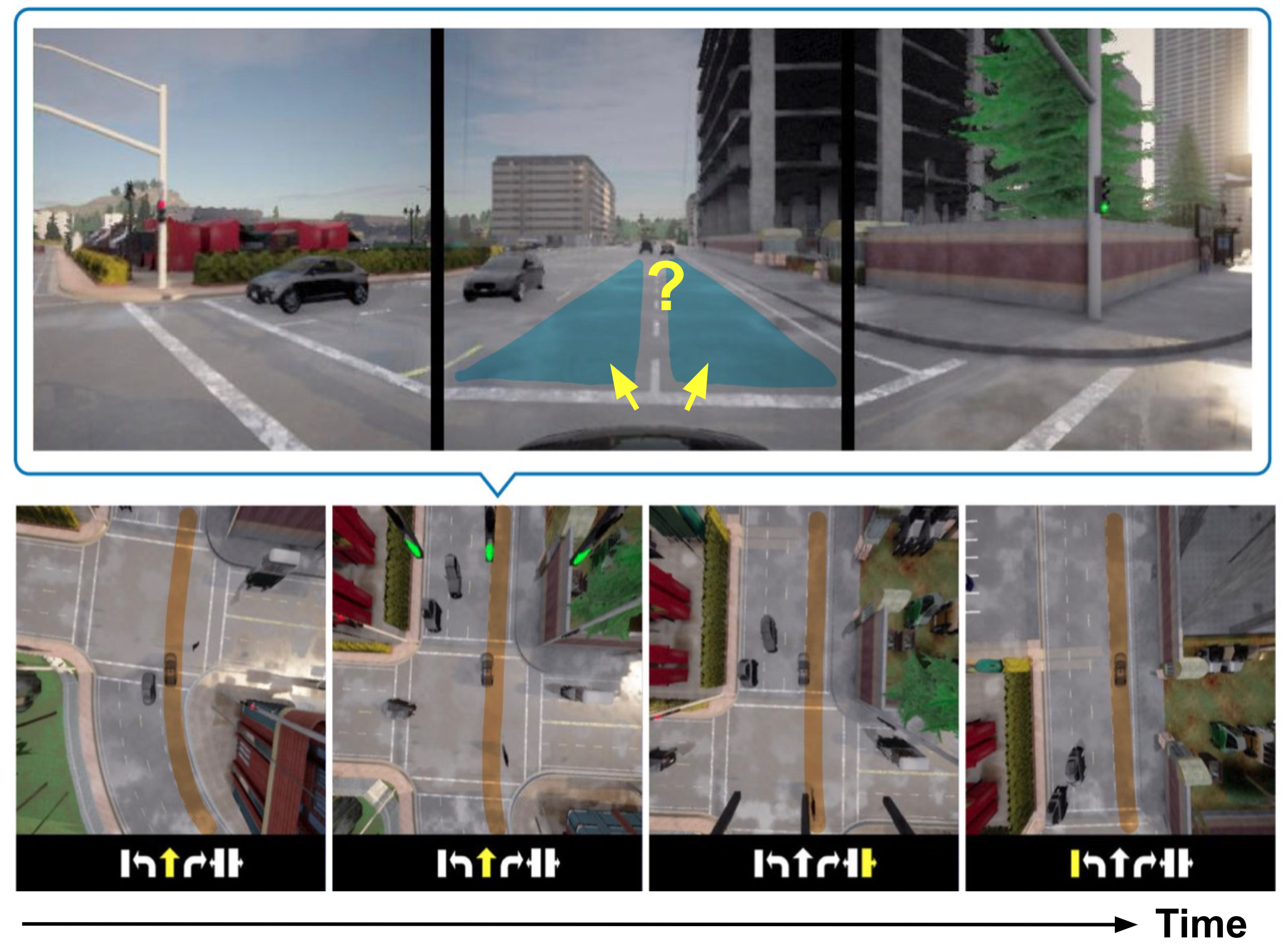}
  \caption{Top: when the ego-vehicle is entering a new road segment from an intersection, the \emph{go-straight} navigation command is ambiguous. The ego-vehicle can legally move to any of the highlighted lanes. Bottom: four aerial views at different times with the pre-planned global trajectory shown in \textbf{\textcolor{orange}{orange}}. They illustrate how the high-level navigation command changes to \emph{move-to-right-lane}, to inform how to get back to the desired trajectory.}
  \label{fig:command_ambiguous}
\end{figure}

\section{EXPERIMENTS}
\subsection{Training Datasets}\label{sec:Dataset}
In order to conduct our experiments, we use the CARLA simulator \cite{Dosovitskiy:2017} 0.9.13, which was the latest official version when we started this research. As recent top-performing methods \cite{Hu:2022}, for on-board data collection in CARLA, we use the \emph{teacher} expert driver from \cite{Zhang:2021}, termed as Roach RL since it is based on reinforcement learning and was trained with privileged information. Roach RL shows a more realistic and diverse behavior than the default (handcrafted) expert driver in CARLA. Note that in real-world experiments we would use different human drivers as experts. We use the default settings of \cite{Zhang:2021}, so as in the \emph{student} driver of \cite{Zhang:2021} (RIM) as well as in \cite{Hu:2022} (MILE), the ego-vehicle is the Lincoln 2017 available in CARLA. Each of our three forward-facing cameras on-board the ego-vehicle has a resolution of $W\times H=300\times300$ pixels, covering an HFOV of $60^{\circ}$. They are placed without overlapping so that they jointly cover an HFOV=$180^{\circ}$ centered in the main axis of the ego-vehicle. 

With such expert driver, ego-vehicle, and on-board cameras, we collect data for increasingly complex experiments. First, we collect a dataset from Town01 in CARLA, which is a small town only enabling single-lane driving, {\ie}, lane change maneuvers are not possible. In particular, we collect 15 hours of data at 10 fps ($\sim$540K frames from each camera view), under 4 training weathers, namely, ClearNoon, ClearSunset, HardRainNoon, and WetNoon. In this case, CARLA's Town02 is used for generalization testing under SoftRainSunset and WetSunset weather conditions. Second, we collect a dataset from multiple CARLA's towns to include more complex scenarios such as multi-lane driving, entering and exiting highways, passing crossroads, {\etc} In order to keep the same setting as \cite{Hu:2022}, we hold Town05 for testing, and collect 25 hours of data at 10 fps from Town01 to Town06 (5 hours per town; $\sim$900K frames from each camera). Training and testing weathers are the same for both Town01 and Town02. 

\subsection{Training Details}\label{sec:optparam} 
To optimize Eq. (\ref{eq:loss}), we use the Adam \cite{Kingma:2015} with an initial learning rate of $10^{-4}$ and weight decay of $0.01$. We train for 80 epochs on 2 NVIDIA A40 GPUs in parallel, with a batch size of 120. The learning rate decays by half at epochs 30, 50, and 65. 

\begin{figure}
  \centering
     \begin{subfigure}[b]{\linewidth}
         \centering
         \includegraphics[width=\textwidth]{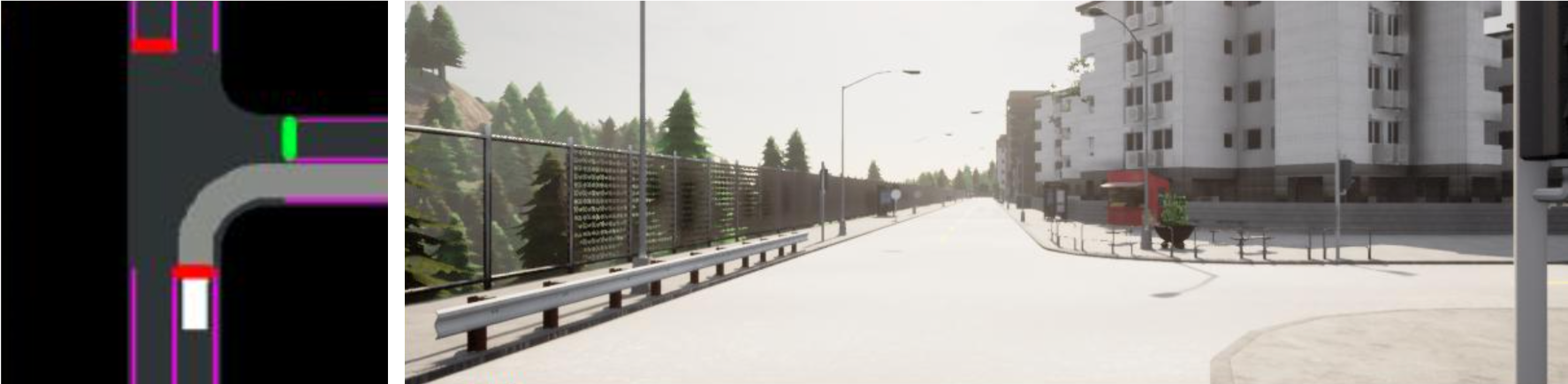}
         \caption{Roach RL \cite{Zhang:2021}: using semantic BeV as input during training time}
         \label{fig:TLCausualConfusionB}
     \end{subfigure}
     \hfill
     \begin{subfigure}[b]{\linewidth}
         \centering
         \includegraphics[width=\textwidth]{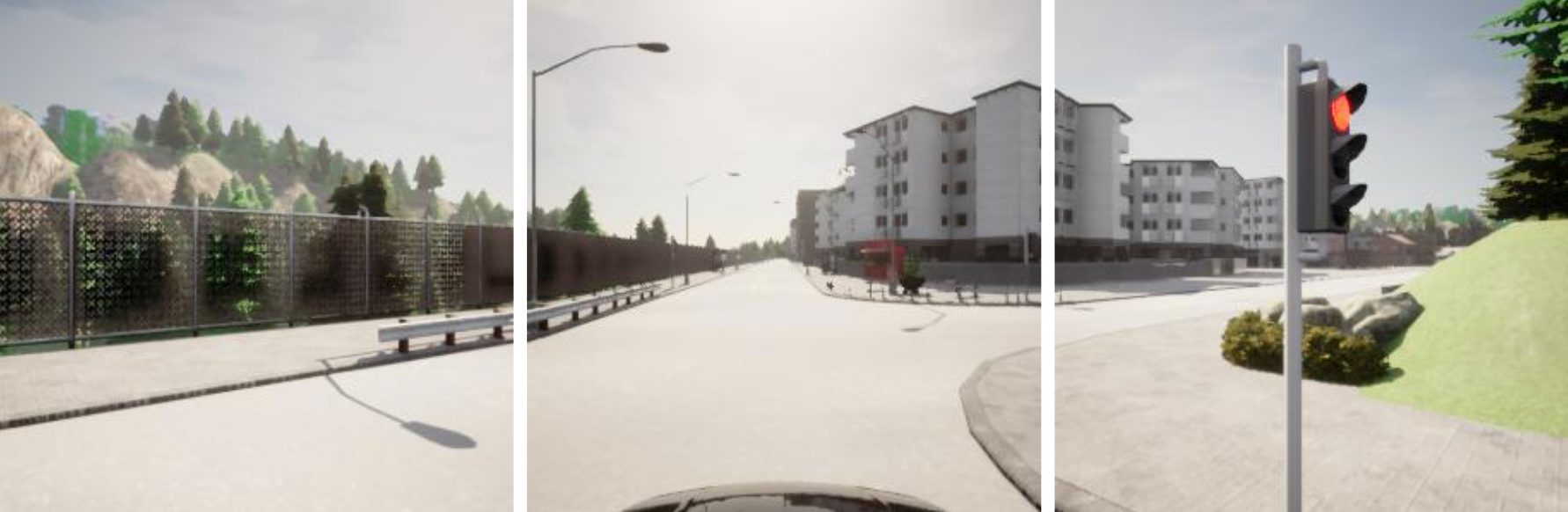}
         \caption{CIL++: only images from three views as input}
         \label{fig:TLCausualConfusionA}
     \end{subfigure}
  \caption{Top Left: The expert we use for data collection is the teacher agent Roach RL \cite{Zhang:2021}, which has access to semantic BeVs. Note that, in simulation, this expert plays the role of a human driver in the real world, who would drive to collect on-board data. Top Right: Using an insufficient FOV, the red traffic light is not observable in the image when the expert stops close to it, which may cause causal confusion when applying imitation learning to train the student agent RIM. Bottom: CIL++ avoids this causal confusion by using a larger HFOV based on three complementary images (from different cameras). More specifically, RIM uses HFOV=$100^{\circ}$, while CIL++ uses HFOV=$180^{\circ}$.}
  \label{fig:causalconfusion}
\end{figure}

\subsection{Driving Evaluation}
\label{sec:Metrics}

We follow the \emph{NoCrash} benchmark \cite{Codevilla:2019} and the offline CARLA leaderboard benchmark \cite{Zhang:2021,Hu:2022} for experiments on small single-lane towns (Sec. \ref{sec:small_town_results}) and multiple towns (Sec. \ref{sec:multi_towns_result}), respectively.

\subsubsection{NoCrash Metrics.}\label{nocrash_metrics}
It consists of three tasks with increasing levels of difficulty: \emph{Empty}, \emph{Regular}, and \emph{Dense}, according to the number of dynamic objects in the scene ({\ie}, pedestrians and vehicles). In Town 2, the numbers of dynamic objects are specified as:
\begin{itemize}
    \item \emph{Empty}: 0 pedestrian and 0 vehicle;
    \item \emph{Regular}: 50 pedestrians and 15 vehicles;
    \item \emph{Dense}: 150 pedestrians and 70 vehicles
\end{itemize}
In the \emph{Dense} case, the default traffic density set in NoCrash always leads to congestion deadlocks at intersections \cite{Zhang:2021}. Thus, we follow the \emph{Busy} case redefined in \cite{Zhang:2021}, decreasing the number of pedestrians from 150 to 70. Each task corresponds to 25 goal-directed episodes under 2 new kinds of weather. The episode will be terminated and counted as a failure once a collision occurs. For the other infractions, the driving score will be deduced according to the penalty rule in NoCrash. 

The main metric to compare driving models is the success rate (\emph{SR}), which is the percentage of episodes successfully completed. For a fine-grained comparison, in addition, we provide the strict success rate (\emph{S.SR}). It reflects the percentage of successful episodes under zero tolerance for any traffic infraction, such as failing to stop at a red traffic light, route deviation, {\etc}. As a complement, we also include additional infraction metrics: \emph{T.L} is the number of times not stopping at a red traffic light; \emph{C.V} is the number of collisions with other vehicles; \emph{R.Dev} is the number of route deviations, {\ie}, when the high-level command is not well-executed; \emph{O.L} accounts for the ego-vehicle driving out-of-lane ({\eg}, in the opposite lane or in the sidewalk); \emph{C.L} is the number of collisions with the town layout. All infraction values are normalized per driven kilometer.

\subsubsection{Offline Leaderboard Metrics.}\label{lb_metrics}
To align our evaluation with \cite{Hu:2022}, we use the offline CARLA's Leaderboard metrics for multiple towns. The most important metrics are the average driving score (\emph{Avg.DS}) and the average route completion (\emph{Avg.RC}). \emph{Avg.DS} is based on penalizing driving performance according to the terms defined in CARLA's Leaderboard, while \emph{Avg.RC} is the average distance towards the goal that the ego-vehicle is able to travel.

\subsubsection{High-level Navigation Commands} 
As in CILRS \cite{Codevilla:2019}, at training time we use simple navigation commands such as \emph{continue} in the lane, or \emph{go-straight/turn-left/turn-right} next time an intersection is reached. However, in complex towns, after crossing an intersection in any direction, we may legally enter any of the multiple lanes. Thus, since this can be known by the global navigation system, when the ego-vehicle enters a lane out of the pre-planned global trajectory, a corrective command is forced, like \emph{move-to-left-lane} or  \emph{move-to-right-lane} as soon as possible. This corrective mechanism is used only at testing time. Figure \ref{fig:command_ambiguous} provides an example.

\begin{table*}
  \centering
  \resizebox{0.95\linewidth}{!}{
  \begin{tabular}{@{}lccc|ccc|ccc@{}}
    \toprule
          & \multicolumn{3}{c}{Empty} & \multicolumn{3}{c}{Regular} & \multicolumn{3}{c}{Busy}  \\
    
    & \textbf{$\uparrow$ SR(\%)} & \textbf{$\uparrow$ S.SR(\%)} & $\downarrow$ T.L  & \textbf{$\uparrow$ SR(\%)} & \textbf{$\uparrow$ S.SR(\%)} & $\downarrow$ T.L & \textbf{$\uparrow$ SR(\%)} & \textbf{$\uparrow$ S.SR(\%)} & $\downarrow$ C.V \\
    \midrule
    
    RIM  & $100\pm0.0$ & $85\pm1.2$ & $66\pm5.0$ & $97\pm2.3$ & $86\pm7.2$ & $66\pm54$ & $81\pm5.0$& $68\pm7.2$ & $63\pm52.7$\\
    CIL++ & $100\pm0.0$ & $100\pm0.0$ & $0\pm0.0$  & $99\pm2.3$ & $97\pm3.1$ & $7\pm7.9$ & $83\pm7.6$ & $77\pm7.6$ & $45\pm21.5$ \\
    \midrule
    Expert & $100\pm0.0$ & $100\pm0.0$ & $0\pm0.0$ & $100\pm0.0$ & $97\pm0.0$ & $13\pm4.6$ & $84\pm2.0$ & $82\pm2.0$ & $37\pm14.1$ \\
    \bottomrule 
  \end{tabular}
  }
  \caption{Town02 NoCrash results. RIM stands for Roach IL and the Expert is Roach RL \cite{Zhang:2021}. All models are tested on CARLA 0.9.13. Mean and standard deviations are computed using three runs with different seeds. For $\uparrow$, the higher the better, while for $\downarrow$ is the opposite.}
  \label{tab:T2_NC_results}
\end{table*}

\begin{table*}
  \centering
  \resizebox{0.95\linewidth}{!}{
  \begin{tabular}{@{}lcccccccccccccccc@{}}
    \toprule
     &  \textbf{$\uparrow$ SR(\%)} &  \textbf{$\uparrow$ S.SR(\%)} &  \textbf{$\uparrow$ Avg.RC(\%)} &  \textbf{$\uparrow$ Avg.DS}  
     & $\downarrow$ C.L & $\downarrow$ T.L & $\downarrow$ O.L & $\downarrow$ R.Dev 
     \\
    \midrule
    HFOV $100^{\circ}$ & $52$ & $40$ & $83$ & $69.1$ 
    &$428.6$& $19.6$ & $ 339.6$ & $10.7$ 
    \\
    HFOV $180^{\circ}$ & $100$ & $98$ & $100$ & $99.2$ 
    &$7.3$& $0.0$ & $0.0$ & $0.0$ 
    \\
    \midrule
    Expert & $100$ & $100$ & $100$ & $100.0$ 
    & $0.0$ & $0.0$ & $0.0$ & $0.0$ 
    \\
    \bottomrule 
  \end{tabular}
  }
  \caption{Impact of sensor suite HFOV in the NoCrash Regular case. For HFOV=$100^{\circ}$ we use a single camera with a resolution of $W\times H=600\times170$ pixels, while for HFOV=$180^{\circ}$ we use the multi-view setting detailed in the main text.}
  \label{tab:FOVresults}
\end{table*}

\begin{table*}
  \centering
  \resizebox{0.95\linewidth}{!}{
  \begin{tabular}{@{}lccccccccccccccccccccc@{}}
    \toprule
           & \textbf{$\uparrow$ Avg.RC(\%)} & \textbf{$\uparrow$ Avg.DS }
          & $\downarrow$ C.V & $\downarrow$ C.L & $\downarrow$ T.L & $\downarrow$ O.L & $\downarrow$ R.Dev 
          \\
    \midrule
    RIM  & $92\pm3.1$ & $51\pm7.9$ 
    & $7.5\pm1.3$ & $4.3\pm1.6$ & $26.0\pm8.9$ & $5.4\pm2.7$ & $3.0\pm3.2$ & \\
    MILE 
    & $98\pm2.2$ & $73\pm2.9$ 
    & $6.0\pm3.7$ & $0.0\pm0.0$ & $3.6\pm3.8$ & $3.5\pm1.5$ & $0.0\pm0.0$ \\   
    CIL++ 
    & $98\pm1.7$ & $68\pm2.7$ 
    & $6.0\pm0.5$ & $3.8\pm0.7$ & $5.8\pm5.1$ & $6.1\pm2.2$ & $9.4\pm3.6$ \\
    \midrule
    Expert 
    & $99\pm0.8$ & $89\pm1.7$ 
    & $3.2\pm1.1$ & $0.0\pm0.0$ & $1.3\pm0.4$ & $0.0\pm0.0$ & $0.0\pm0.0$ \\
    \bottomrule 
  \end{tabular}
  }
  \caption{Town05 results according to CARLA's offline metrics. All models are tested on CARLA 0.9.13. Mean and standard deviations are computed using three runs with different seeds. For $\uparrow$, the higher the better; for $\downarrow$, the opposite.}
  \label{tab:T5_results}
\end{table*}

\begin{figure*}[ht!]
  \centering
  \includegraphics[width=\linewidth]{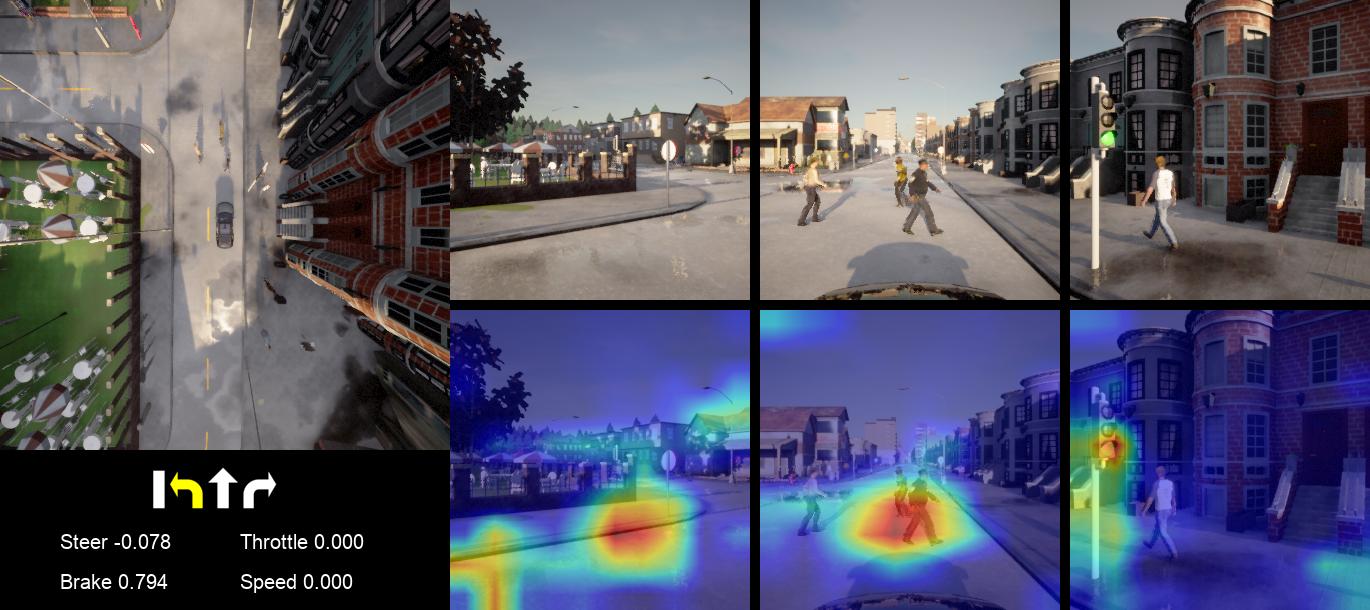}
  \caption{Activation maps of CIL++ at an intersection in Town02. Three image areas from different views are highly activated: the traffic light at the right image, the crossing pedestrians at the central one, and the lane shoulder at the left one. Causality between observation and action is shown as a strong braking (0.794) due to the pedestrians, even though the traffic light in green and the \emph{turn-left} command.}
  \label{fig:attention_ped_greed}
\end{figure*}

\begin{table}
  \centering
  \begin{tabular}{@{}lccccc@{}}
    \toprule
     & $\uparrow$ SR(\%) & $\uparrow$ S.SR(\%) & $\uparrow$ Avg.RC(\%) & $\uparrow$ Avg.DS  \\
     \midrule
    LF.A & $72$ & $62$ & $86$ & $78.1$  \\
    LF.C & $76$ & $66$ & $88$ & $80.6$  \\
    Token & $88$ & $80$ & $93$ & $88.4$  \\
    CIL++ & $88$ & $84$ & $93$ & $88.8$  \\ 
    \bottomrule 
  \end{tabular}
  \caption{Results of different data input fusion approaches for NoCrash Busy scenarios.}
  \label{tab:ablation_study_inputfusion}
\end{table}

\begin{table}
  \centering
  \begin{tabular}{@{}lcccccccccccccccc@{}}
    \toprule
     & $\uparrow$ SR(\%) & $\uparrow$ S.SR(\%) & $\uparrow$ Avg.RC(\%) & $\uparrow$ Avg.DS  \\
     \midrule
    GAP & $64$ & $46$ & $87$ & $75.1$  \\
    VS & $70$ & $62$ & $89$ & $78.6$  \\
    CIL++ & $88$ & $84$ & $93$ & $88.8$  \\ 
    \bottomrule 
  \end{tabular}
  \caption{Results of different multi-view fusion approaches for NoCrash Busy scenarios.}
  \label{tab:ablation_study_sa}
\end{table}


\subsection{Experimental Results}
\label{sec:Results}
We compare CIL++ with two SOTA vision-based EtE-AD models, namely, the Roach IL model (here RIM) \cite{Zhang:2021} and MILE \cite{Hu:2022}. 
Notice that although CIL++ uses the data generated by the Roach RL model, essentially training a CIL++ model does not require human-labeled sensor data. In our case, the Roach RL plays the role of a human driver at data acquisition time. In contrast, training a RIM model requires teaching from the Roach RL expert who was trained with privileged information, while MILE is trained with semantic BeV as supervision.

\subsubsection{Small Single-lane Towns} \label{sec:small_town_results}
We first use CARLA's Town01 and Town02 along with the NoCrash metrics (Sec. \ref{nocrash_metrics}) for initial experiments. Town01 is used for training and Town02 for testing (Sec. \ref{sec:Dataset}). MILE only provides a model trained on CARLA's multiple towns, but there is no model trained only on Town01, while RIM has versions trained on Town01 and multiple towns. Thus, for a fair comparison, we only use RIM's single-town trained model. We show in Table \ref{tab:T2_NC_results} SR and S.SR for the considered traffic densities (Empty, Regular, Busy). In order to have a more focused evaluation, we show T.L only for the Empty and Regular cases, while C.V is shown only for the Busy case. Note that scenarios with no or few dynamic obstacles can better show the ego-vehicle reaction to red traffic lights, while collisions are better evaluated in scenarios with more dynamic objects. 

In general, CIL++ achieves the best results in all the tasks. In the Empty case, CIL++ clearly outperforms RIM in avoiding traffic light infractions, which also contributes to a better S.SR. We reach the same conclusion in the Regular case. In the Busy case, 
CIL++ reaches almost the expert's performance, again, being clearly better than RIM for S.SR and producing fewer collisions with vehicles. For the expert, the failure cases in Busy scenarios are due to traffic deadlocks, which lead to a timeout in route completion. Thus, its performance still can be considered as a proper upper bound. 

Traffic lights tend to be on sidewalks, so detecting them from a close distance requires a sensor setting with a proper HFOV. Otherwise, causal confusion can appear. We think that the poor performance of RIM on the T.L metric is due to a narrow HFOV as illustrated in Fig. \ref{fig:causalconfusion}.  
To confirm this hypothesis, we conduct experiments using two HFOV settings for CIL++, 100 and 180. As seen in Table \ref{tab:FOVresults}, we note that the number of infractions (T.L, C.L, O.L, R.Dev) increase when we use a lower HFOV=$100^\circ$ compared to HFOV=$180^\circ$. For HFOV=$100^\circ$, we have observed that the track of the road shoulder is easily out-of-observation at intersections, leading to more O.L, C.L, and R.Dev. For HFOV=$180^{\circ}$, the ego-vehicle can better perform the right driving maneuver, thanks to having the road shoulder as a reference.

\subsubsection{Multi-town Generalization}\label{sec:multi_towns_result}
In this section, we assess the performance of CIL++ in much more complex scenarios, as provided by CARLA's multiple towns. As mentioned in Sec. \ref{sec:Dataset}, for a fair comparison, we align the training and testing settings with MILE \cite{Hu:2022}, using CARLA's offline Leaderboard metrics (Sec. \ref{lb_metrics}). The results for all models trained on multi-town data are shown in Table \ref{tab:T5_results}. RIM shows the worst performance among the three models, incurring more infractions, thus obtaining a significantly lower Avg.DS. CIL++ achieves 98\% Avg.RC, which is on par with MILE. In terms of Avg.DS, MILE remains the best scoring, yielding a 73\% while CIL++ achieves a 68\%. We observe that this is because MILE seldom drives outside the pre-planned lane, given the route map as input. On the contrary, CIL++ lacks the explicit use of this route map since it only receives high-level navigation commands.

\subsubsection{Visualizing CIL++'s Attention}
\label{sec:Visualization}
We are  interested in the image content to which CIL++ pays attention. Following Grad-CAM \cite{Selvaraju:2017}, gradients flow from the action space to the final convolutional layer of the ResNet backbone. This should produce a map that highlights the important image areas for action prediction. However, since CIL++ solves a regression task, its output could be either negative or positive values, while Grad-CAM is originally designed for image classification tasks which always provide positive outputs. To adapt Grad-CAM to our case, we cannot merely take into account the positive gradient of the feature map. The computation should be divided into two cases depending on the sign of the output value. Negative gradients are used to calculate the weights for the feature map when the acceleration or steering angle value is lower than zero, otherwise, the positive gradient is used.

Fig. \ref{fig:attention_ped_greed} shows the activation map at an intersection. Three image areas are highly activated: the traffic light in the right image, the crossing pedestrians in the central one, and the lane shoulder in the left one. Thus, we believe that CIL++ shows a proper understanding of this scene, and a clear causality between observation and action since it decides to brake due to the pedestrians, even though the traffic light is in green and a \emph{turn-left} navigation command is given.

\subsubsection{Ablation Study}
\label{sec:Ablation Study}
To inspect the impact of some components of CIL++, we provide an ablation study. Specifically, we are interested in the fusions of input data and multi-view. 

\paragraph{Input Data Fusion}
EtE-AD models require not only sensor data but also signal information, like the ego-vehicle speed and a high-level navigation command. It is interesting to study how to properly fuse these inputs. To compare, we implement several types of input data fusion in Table \ref{tab:ablation_study_inputfusion}: adding, concatenation, and tokenization. In the first, the speed and command features are simply added to the image features. This addition could be done either before (the default operation in CIL++) or after the Transformer Encoder block. We name the latter as late fusion adding (LF.A). Another common data fusion method is concatenation, which firstly stacks all the features and takes an extra join FC layer to fuse them, which we term as late fusion concatenation (LF.C). Since the transformer model uses a self-attention mechanism to fuse features between tokens, we can tokenize the speed and navigation command features and feed them into the transformer block along with the image features. We term this approach as Token. Our results suggest that there is no obvious difference between tokenization and early adding fusion. These two approaches show better results than the late fusion.

\paragraph{Multi-view Fusion}
CIL++ uses attention layers to fuse multi-view information. To understand their contribution, we remove the transformer block and simply use the ResNet34 which retains the average pooling and an FC layer for embedding each image view. The embedding outputs are then stacked and fed to the FC join layers for fusion. We term this approach as view stacking (VS) in Table \ref{tab:ablation_study_sa}. The speed and command features are added to the joint embedding before feeding into the action prediction MLP. We see that without the self-attention layers, the SR drops from 88\% to 70\%. We think this is because the average pooling layer causes a loss of spatial information, while this information is very important for actual driving. The agent should take different actions according to the location of dynamic obstacles. We also use a transformer block to process the output of the ResNet average pooling layer (GAP), instead of using the flattened feature map from the last convolutional layer of ResNet34. The results drop significantly, {\eg}, and the SR goes from 88\% to 64\%.

\section{CONCLUSIONS}
We have presented CIL++, which aims at becoming a new strong baseline representing vision-based pure EtE-AD models trained by imitation learning. This is required because recent literature may lead to the conclusion that such approaches are poorly performing compared to those relying on additional and costly supervision. We have argued that previous vision-based pure EtE-AD models were developed in sub-optimal conditions. Thus, we have developed a model which relies on three cameras (views) to reach an HFOV=$180^\circ$ and a more realistic expert driver to collect on-board data in the CARLA simulator. We have proposed a visual transformer that acts as a mid-level attention mechanism for these views, allowing CIL++ to associate feature map patches (tokens) across different views. CIL++ performs at the expert level on NoCrash metrics and is a pure EtE-AD model capable of obtaining competitive results in complex towns. We have presented an ablative study showing the relevance of all the components of CIL++.
In future work, we plan to add rear-view cameras, to improve lane change maneuvers.



\section*{ACKNOWLEDGMENT}
This research is supported as a part of the project TED2021-132802B-I00 funded by MCIN/AEI/10.13039/501100011033 and the European Union NextGenerationEU/PRTR. Yi Xiao acknowledges the support to her PhD studies provided by the Chinese Scholarship Council (CSC), Grant No.201808390010. Diego Porres acknowledges the support to his PhD studies provided by Grant PRE2018-083417 funded by MCIN/AEI /10.13039/501100011033 and FSE invierte en tu futuro. Antonio M. López acknowledges the financial support to his general research activities given by ICREA under the ICREA Academia Program. Antonio thanks the synergies, in terms of research ideas, arising from the project PID2020-115734RB-C21 funded by MCIN/AEI/10.13039/501100011033. The authors acknowledge the support of the Generalitat de Catalunya CERCA Program and its ACCIO agency to CVC’s general activities.


\bibliographystyle{IEEEtran}
\bibliography{reference}

\end{document}